\documentclass[11pt]{article}

\usepackage[preprint]{acl}

\usepackage{times}
\usepackage{latexsym}

\usepackage[T1]{fontenc}
\usepackage[utf8]{inputenc}

\usepackage{microtype}

\usepackage{inconsolata}

\usepackage{graphicx}

\usepackage{booktabs}
\usepackage{amsmath}
\usepackage{amssymb}
\usepackage{tikz}
\usetikzlibrary{positioning}

\title{Internalizer: Portable Context-to-Parameter Mapping for Very Large Language Models}

\author{
  Peter A. Devine, Nick Ryan, Benjamin Sirb, Alex Chiocchi \\
  Interval \\
  \texttt{\{peter.devine, nick.ryan, ben, alex\}@ai-interval.com}
}

\begin{document}
\maketitle
\begin{abstract}

Hypernetworks that map a context directly to a LoRA adapter let a
large language model carry that context in its weights, but prior work
has demonstrated them only on base models of up to 14 billion parameters.\\
We present the Internalizer, a state-of-the-art, portable
Context-to-Parameter Mapping hypernetwork that generates
document-specific LoRA adapters for the frozen 284B-parameter DeepSeek
v4 Flash, a target twenty times larger than in any previous
work. Most of its parameters
live in a model-agnostic trunk with only thin entry and exit layers
per base model, so it trains cheaply against small models before
being ported to the large one.\\
On unseen documents of up to 4096 tokens, the generated adapters reach 84.9\%
top-1 and 97.8\% top-5 teacher-forced accuracy against 63.4\% and
83.5\% for the base model, with nothing in the context window but a
three-word instruction.\\
Once the hypernetwork is trained, a single forward pass turns any
document into an adapter for such a model, which could be served alone
for speed or alongside the document in the window to raise accuracy
further.

\end{abstract}

\section{Introduction}
\label{sec:intro}

Large Language Models (LLMs) have become widely used in commercial and
information-technology pipelines~\citep{bick2024rapid}, largely because they offer higher
accuracy and lower cost than human labour in many areas~\citep{gilardi2023chatgpt}.

However, these advantages do not extend to all tasks, as for many use
cases LLM inference remains too expensive~\citep{zhou2024survey}, or model accuracy too
low~\citep{huang2025hallucination}, to justify deployment.

Three major causes of high cost and low accuracy are bloated context
windows~\citep{jiang2023llmlingua}, the inefficiency of in-context learning~\citep{brown2020language,mu2023gist}, and the
impracticality of per-task training~\citep{hu2022lora}.

Commercial tasks often fill the context window with task-relevant
information, raising the latency, and thus the cost, of inference, while the
"lost-in-the-middle" phenomenon erodes accuracy as context length
grows~\citep{liu2024lost}.

In-context learning of complex tasks requires task explanations and worked
examples in the window, and often long chain-of-thought traces before an
output, driving up cost further.

Fine-tuning primes a model for a specific task, but it is too expensive and
slow for all but the most important repeated tasks~\citep{han2024peft}. Even then, the
resulting task adapter knows nothing of the particular input it will be
used on.

To tackle these three challenges, previous work has proposed
Context-to-Parameter Mapping, a hypernetwork framework that generates a
context-specific LoRA adapter for a given task~\citep{gao2026sada}. The context can be a
document, a code file, a set of table schemas, an image, or any other
tokenizable LLM input. The hypernetwork converts this context into a LoRA
adapter, and at inference time the model input, a user question, command,
or instruction relating to that context, produces the model output as in
ordinary LLM use.

\begin{figure}[h]
    \centering
    \makebox[\columnwidth][c]{\includegraphics[width=1.1\columnwidth]{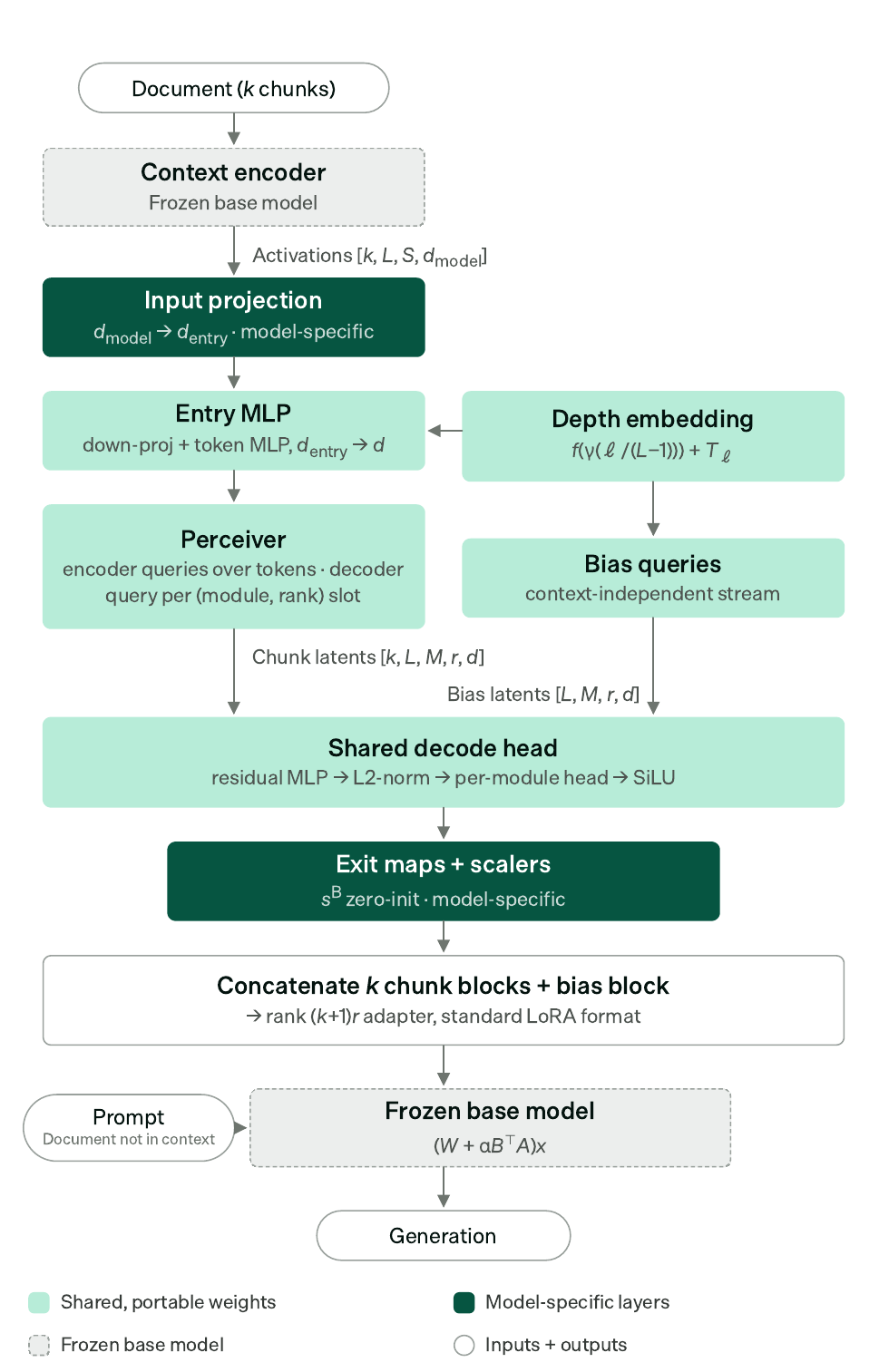}}
    \caption{System architecture, with the shared trunk in light green and
    the thin model-specific entry and exit layers in dark green.}
    \label{fig:architecture}
\end{figure}

The adapter is computed once per context and can be pre-generated and
cached, effectively internalizing the context within the LLM. Since the
hypernetwork is meta-learned, this adaptation is cheap and near-instant
even for very large language models, in contrast to per-task fine-tuning.
The adapter can then be served in two ways. In a speed mode it
replaces the context in the window, reducing inference time and cost, or
in an accuracy mode the context stays in the window and the adapter is
served alongside it, which is the setting we expect to matter most in
practice.

Accuracy improves because the adapters are not only specialised for a
given task (e.g. document QA, code generation, table traversal, image
QA) but also carry knowledge of the specific context they will be used
on. This is, in effect, a superset of standard LoRA training, with the
adapter conditioned on the context as well as the task. Previous work has shown that such adapters can outperform in-context
use of the very context they were generated
from~\citep{yu2026latentskill}, and an adapter served alongside that
context has both sources to draw on.

However, Context-to-Parameter Mapping has only ever been demonstrated on small models and
academic benchmarks, and no prior work has targeted a language model beyond
14 billion parameters. This limits its applicability to commercial tasks,
which often require the accuracy of far larger models.

We present the Internalizer, a state-of-the-art, portable
Context-to-Parameter Mapping hypernetwork that scales to LLMs with
hundreds of billions of parameters. Its largely model-agnostic architecture allows the
hypernetwork to be re-used across LLMs, massively reducing re-training cost,
and enables application to much larger language models than was previously
feasible.

Four main design choices carry this result. Oversized entry and exit
projections decouple the trunk's representational capacity from any
target model's hidden width. Adapter slots are restricted to the gated
MLP projections that virtually every modern backbone shares, which is
what lets one trunk serve many base models. An error-gated loss focuses
training on the hard tokens, such as names and numbers, that downstream
tasks depend on. Finally, a strategy we call reverse distillation does
the bulk of the learning cheaply against small base models before
porting to the expensive target.

In our experiments, we internalize documents up to 4096 tokens long into a
284B-parameter model (DeepSeek v4 Flash 0731~\citep{deepseekv4flash}) at 84.9\% top-1
teacher-forced accuracy against 63.4\% for the base model, demonstrating
that this approach is tractable even for some of the largest
state-of-the-art LLMs. This is a target model twenty times
larger than in any previous research~\citep{muhtar2024streamadapter},
which to our knowledge makes the Internalizer the state-of-the-art
hypernetwork for context internalization.

This makes Context-to-Parameter Mapping much more practically useful. 
At the scale of prior work, models are small enough
that their accuracy often falls short of commercial requirements, so a
context-specific adapter has little to build on. At the scale of
state-of-the-art models the base accuracy is high enough that a further
gain is worth having, and training an individual LoRA per task or
context grows more expensive with model size, exactly where a single
hypernetwork forward pass replaces it. The training cost is paid once,
and from then on any document, schema, or code file becomes a
context-informed LoRA adapter on demand, whether served alone in speed
mode or in accuracy mode alongside the context it was generated from.
The Internalizer is fully
model-agnostic, and we aim to apply it to improve accuracy, and where
the context can be dropped to reduce inference cost, across a variety
of real-world use cases.

This paper sets out our architecture, pre-training strategy, and vision for
how this technology can be used.

\section{Related work}
\label{sec:relatedwork}

Context-to-Parameter Mapping~\citep{gao2026sada} trains a
hypernetwork~\citep{ha2016hypernetworks} to convert a context directly into
a weight update for a frozen LLM, so that the context no longer needs to
occupy the context window.

Two early systems realised this idea in late 2024.
GenerativeAdapter~\citep{chen2025generativeadapter} reads the frozen
model's hidden states as context arrives and accumulates them into a
LoRA adapter, internalizing documents, demonstrations, and user
conversations at a fraction of the cost of prompting.
StreamAdapter~\citep{muhtar2024streamadapter} instead compresses the KV
cache of tokens evicted from the context window, matching or beating
in-context learning with far fewer demonstrations and reaching the
largest target LLM to date, the 14B Phi-3-Medium~\citep{abdin2024phi3}. Both, however, pack the context into an
adapter of fixed capacity however long it grows (ours gains rank with
every additional chunk), and neither hypernetwork can be reused with
another base model.

Doc-to-LoRA (D2L)~\citep{charakorn2026doctolora} subsequently reworked
the architecture of the hypernetwork itself, basing it on a Perceiver
encoder that compresses each chunk of base-model activations into
latents that decode to a LoRA block. Blocks concatenate along the rank
dimension so that adapter capacity grows with context length rather
than staying fixed, and training minimises the KL divergence to the
base model's own context-conditioned responses. We adopt this chunked
Perceiver design as our starting point. However, D2L's self-distillation
objective means a weak teacher controls what is internalized, its
document-QA training data narrows it to that one task, each new base
model requires retraining from scratch, and it was only demonstrated on
2--7B targets.

Work in 2026 then specialised and diagnosed the approach.
LatentSkill~\citep{yu2026latentskill} internalized SKILL.md files, the
skill documents that agent frameworks otherwise paste into the prompt at
every step, outperforming in-context use of the same files. It serves only
this one task, however, and targets a single 8B backbone.

SADA~\citep{gao2026sada} then investigated why mapped models fail, identifying
hidden-state collapse, where incorrect answers coincide with deep-layer
divergence from the full-context oracle. Its ablations, showing that
supervision quality drives internalization fidelity and that accuracy
rises monotonically with LoRA rank, independently motivate two of our
choices, ground-truth teacher-forced supervision and
context-proportional rank. It too, however, trains one hypernetwork per
base model and evaluates only 1.5--7B backbones.

Closest to our portability goal, PorTAL~\citep{geist2026portal} showed
that a task adaptation can be learned once and moved cheaply across base
models. A shared decoder core generates per-layer LoRA from a
base-agnostic task latent, and porting to an unseen model refits only a
thin per-base alignment, recovering ${\sim}98\%$ of a from-scratch
LoRA's lift. This independently validates the factorisation we rely on,
a large shared core behind thin per-model maps. Each task, however, is
a single learned latent with no context input, so every new task
requires its own training run rather than a forward pass, the adapter
carries the task but nothing of the input it is applied to, and its
bases span only 1.7--8B parameters.

Our training strategy also inverts the usual direction of knowledge
distillation~\citep{hinton2015distilling}, which transfers capability
from a large, expensive teacher into a small student. Small-to-large
transfer has precedent in other settings~\citep{burns2023weaktostrong} 
and the term reverse distillation itself appears in anomaly detection,
where a student reconstructs its teacher's representations in the
reverse direction~\citep{deng2022reverse}. No work, however, warms up a
hypernetwork on small base models before porting it to a much larger one. To
our knowledge this inversion is novel for context-to-parameter mapping,
and we call the strategy reverse distillation, in deliberate contrast
to the conventional direction.

Previous work has provided a promising mechanism for internalizing context
within model parameters, but two constraints remain. Every
context-internalizing hypernetwork is bound to the single base model it
was trained on, and no
target LLM has exceeded ${\sim}$14B parameters. The Internalizer addresses
both by placing its heavy computation in a shared, model-agnostic trunk
with thin per-model entry and exit layers (${\lesssim}10\%$ of parameters),
making porting cheap, and by internalizing contexts into a 284B-parameter
target~\citep{deepseekv4flash}, twenty times the scale of prior work
and the regime where base accuracy is high enough for a
context-specific adapter to be commercially useful. In target scale and in portability, the Internalizer therefore
advances the state of the art for this class of hypernetwork.

\section{Architecture}
\label{sec:arch}

This section describes the architecture of the Internalizer hypernetwork and
the design decisions behind it.

At a high level, the Internalizer is a network that generates a LoRA adapter
for a base LLM. Given a context, it reads the per-layer activations produced
by running that context through a frozen base model, and produces an adapter
that folds its content into the weights of every layer. The base model itself is never modified
or fine-tuned as the Internalizer does all the work of turning context into
weights. Most of its parameters are shared independently of base models, with
only thin input and output layers being model-specific, so the same
hypernetwork can be ported to new models cheaply.
This architecture is illustrated in Fig~\ref{fig:architecture}.

The Internalizer has six main parts.

\begin{itemize}
  \item \textbf{A hidden state input adapter.} Projects the base model's
    per-layer activations into the trunk's shared latent space.
  \item \textbf{A depth embedding.} A smooth encoding of each layer's
    relative depth, shared across base models.
  \item \textbf{A perceiver.} Compresses each (\texttt{layer}, \texttt{chunk})'s
    variable-length token sequence into a fixed set of latent vectors.
  \item \textbf{Bias queries.} A second stream that generates the
    context-independent part of the adapter.
  \item \textbf{A shared decode head.} Turns each latent vector into the
    $A$ and $B$ rows of one LoRA rank slot.
  \item \textbf{A combine step.} Concatenates chunk and bias blocks along
    the rank axis into one standard LoRA adapter.
\end{itemize}

\subsection{A hidden state input adapter}

We want the hypernetwork to be as model-agnostic as possible, yet every
base model has its own hidden size. The input adapter therefore
takes each model's per-layer token activations and projects them into a
fixed-size latent space that the rest of the hypernetwork operates in. We do
this in two steps: a small model-specific projection first maps the
activations up into an intentionally large intermediate space (called the
entry size, 8192 in our system), and a shared, portable projection then
compresses them down to the latent size (1536), the width every later
stage of the trunk operates at. Because the entry size is larger than any model's
hidden size, the model-specific step is not forced to discard information. All of
the difficult, lossy compression can instead be done by shared weights that
are reused across base models. This oversized intermediate state is
an important part of the architecture. It decouples the
trunk's representational capacity from the hidden width of any
particular base model, so the same trunk can serve targets of any size,
and the decode head below mirrors the same principle at the exit. Small MLPs with non-linearities add expressivity, and a depth
signal is added to each token so later stages know which layer it came from.
For a token activation $h$ from layer $\ell$ we compute
\begin{equation}
\begin{aligned}
x &= W_{\text{down}}\,\operatorname{SiLU}\!\big(W_{\text{proj}}\,
  \operatorname{rmsnorm}(h)\big) + e_\ell, \\
z &= x + \operatorname{MLP}\!\big(\operatorname{rmsnorm}(x)\big),
\end{aligned}
\end{equation}
where $d_{\text{model}}$ is the base model's hidden size, $d$ is the trunk's
latent size,
$W_{\text{proj}} \in \mathbb{R}^{d_{\text{entry}} \times d_{\text{model}}}$
is the model-specific up-projection,
$W_{\text{down}} \in \mathbb{R}^{d \times d_{\text{entry}}}$ and the residual
token MLP are shared, and $e_\ell$ is the depth embedding below. The adapted
tokens $z$ are what the perceiver attends over.

\subsection{A depth embedding}

The hypernetwork generates a different weight update for every layer of the
base model, so it needs to know which layer it is working on. Rather than
using the raw layer index, we encode each layer's relative depth,
how far through the model it sits, from 0 to 1. This way, a 24-layer model
and a 93-layer model use the same embedding in the same way, which is
essential for sharing the hypernetwork across models. The embedding is
smooth, cheap to compute, and independent of the context. It is
computed as
\begin{equation}
\begin{aligned}
e_\ell &= f\big(\gamma(t_\ell)\big) + T_\ell, \qquad
t_\ell = \frac{\ell}{L-1}, \\
\gamma(t) &= \big[\sin(2^k \pi t),\ \cos(2^k \pi t)\big]_{k=0}^{F-1},
\end{aligned}
\end{equation}
where $\ell$ is the layer index, $L$ is the base model's number of layers,
$t_\ell$ is the resulting relative depth, $\gamma$ maps it to Fourier
features at $F{=}8$ frequencies, $f$ is a small shared MLP over those
features,
and $T$ is a zero-initialised model-specific per-layer table that absorbs
each model's absolute layer quirks. The smooth part $f(\gamma(t_\ell))$ is
what is shared across base models.

\subsection{A perceiver}

Contexts vary in length, but the adapter we generate must always have the
same shape, and the perceiver resolves this mismatch. A fixed set of learned
queries (256 in our system) attends over each chunk's tokens, distilling
however many tokens there are into that fixed number of latent vectors. A decoder stage then
cross-attends over these with one query per (\texttt{module}, \texttt{rank}) slot of a single
chunk's rank-$r$ block, the same queries being reused for every chunk,
\begin{equation}
\begin{aligned}
Z_{\ell,c} &= \operatorname{Enc}\big(Q_{\text{enc}},\ z_{\ell,c}\big), \\
U_{\ell,c} &= \operatorname{Dec}\big(Q_{\text{dec}},\ Z_{\ell,c}\big)
  \in \mathbb{R}^{Mr \times d},
\end{aligned}
\end{equation}
where $z_{\ell,c}$ are the adapted tokens of chunk $c$ at layer $\ell$ from
the input adapter, $Q_{\text{enc}}$ and $Q_{\text{dec}}$ are the learned
queries of the two stages, $\operatorname{Enc}$ is blocks of cross-attention
(queries over the chunk's tokens) interleaved with self-attention (nine
blocks in our system, each pairing one cross-attention layer with one
self-attention layer), and $\operatorname{Dec}$ is one cross-attention
block. The output $U_{\ell,c}$ holds one latent vector per slot, with $M$
the number of target modules and $r{=}16$ the per-chunk LoRA rank (48
decoder queries in total), and is handed to the decode head below.
Each (\texttt{layer}, \texttt{chunk}) group is compressed independently, which
lets chunks be processed in parallel and recombined freely later. The
perceiver is also where we deliberately put most of the hypernetwork's
parameters. Its size is chosen so that the model-specific parts never exceed
roughly $10\%$ of the whole system, which is what makes the
hypernetwork portable.

\subsection{Bias queries}

Some of what the adapter needs to express has nothing to do with the
context but is instead a general correction that every adapter should
carry. If each chunk
had to encode this shared component, it would be duplicated $k$ times when a
context's $k$ chunks are combined, growing with chunk count and making chunks
hard to compose. Instead, a second stream of learned queries generates this
context-independent component once, as its own block that is always appended
to the adapter. Chunk blocks then stay purely context-specific and can be
freely combined. The bias latents are built directly from the queries and
depth signal, with no tokens or attention involved,
\begin{equation}
u^{\text{bias}}_{\ell,(m,\rho)} = q_{(m,\rho)} + f\big(\gamma(t_\ell)\big)
  + T^{\text{bias}}_\ell,
\end{equation}
where $q_{(m,\rho)}$ is a learned query for module $m$ and rank slot $\rho$,
$f$ and $\gamma$ are the shared depth encoding of the depth embedding above,
and $T^{\text{bias}}$ is this stream's own zero-initialised model-specific
per-layer table. These latents are decoded by the same head as the chunk
latents. The bias stream is not task-invariant, as we expect it to carry
task-dependent content that should not scale with chunk count, and only
the tiny $T^{\text{bias}}$ table is specific to a base model.

\subsection{A shared decode head}

The decode head turns the latent vectors (from both the chunk stream and
the bias stream) into actual LoRA weight rows. Each vector becomes one
rank slot of the adapter. Mirroring the input adapter, decoding happens at an
intentionally large intermediate size (called the exit size, 2048 in our
system). The expressive,
non-linear work is done there by shared weights, and only thin final
projections into each model's specific weight shapes are model-specific. For
a latent $u$ (a row of $U_{\ell,c}$ from the chunk stream, or a bias latent
$u^{\text{bias}}$), module $m$, layer $\ell$, and rank slot $\rho$, the head
computes
\begin{equation}
\begin{aligned}
\hat u &= \frac{g(u)}{\lVert g(u)\rVert_2}, \qquad
[\,a \,\|\, b\,] = W^{(m)}\hat u, \\
A_{\ell,\rho} &= s^{A}_{m,\ell,\rho}\, E_{d_{\text{in}}}\!
  \operatorname{SiLU}(a), \\
B_{\ell,\rho} &= s^{B}_{m,\ell,\rho}\, E_{d_{\text{out}}}\!
  \operatorname{SiLU}(b),
\end{aligned}
\end{equation}
where $g$ is a shared pre-head residual MLP (a single residual block in our
system),
$W^{(m)} \in \mathbb{R}^{2d_{\text{exit}} \times d}$ is the shared per-module
head whose output splits into the halves $a$ and $b$, and the
$E \in \mathbb{R}^{d_{\text{in/out}} \times d_{\text{exit}}}$ are the
model-specific exit maps (one per distinct row dimension, shared across
modules and $A$/$B$ sides). The rows $A_{\ell,\rho}$ and $B_{\ell,\rho}$ fill
rank slot $\rho$ of the adapter at layer $\ell$. The $s^A, s^B$ are
per-layer, per-slot scalers, with the chunk and bias streams each carrying
their own tables. The $s^B$ of both streams are initialised to zero, so at
the start of training the generated update is exactly zero, the base model
behaves as if unmodified, and training begins from a known-good starting
point.

\subsection{A combine step}

Finally, the pieces are assembled into one adapter per context. The blocks
from its $k$ chunks and the bias block are simply concatenated, giving an
adapter of rank $r(k{+}1)$ in standard LoRA format. Per module and layer,
\begin{equation}
\begin{aligned}
A &= \big[A^{(1)};\ \dots;\ A^{(k)};\ A^{\text{bias}}\big]
  \in \mathbb{R}^{r(k+1) \times d_{\text{in}}}, \\
B &\text{ likewise}, \qquad
\Delta W = \alpha\, B^{\top} A,
\end{aligned}
\end{equation}
so the modulated layer computes $(W + \Delta W)\,x$ on its input $x$, where
$A^{(c)}$ and $B^{(c)}$ are the decoded blocks of chunk $c$,
$A^{\text{bias}}$ and $B^{\text{bias}}$ come from the bias stream, and
$\alpha$ is a fixed scalar hyperparameter. Note that $\alpha$ multiplies the
update directly rather than through the $\alpha/\text{rank}$ convention of
standard LoRA implementations, so the update's scale does not shrink as
chunks are added, requiring setting its scaling to match at inference. 
Because chunks are encoded independently and the shared component
lives in the bias block, combining requires nothing more than concatenation.
We have found this a useful way to scale context length.
Each extra chunk of the context adds its own parameters to the adapter,
so internalizing more tokens naturally means a slightly larger adapter.

\subsection{Targeted modules}
\label{sec:targets}

We target the three MLP projections of every layer, up, gate, and down
($M{=}3$), where D2L~\citep{charakorn2026doctolora} targeted only the down
projections. We target no attention modules. Attention configurations vary
widely across models, while virtually every modern backbone shares the
same gated-MLP structure, so restricting the slots to the MLP is what lets
the same hypernetwork port across base models. On our smaller dense
models these slots are simply each layer's \texttt{up\_proj},
\texttt{gate\_proj}, and \texttt{down\_proj}. On DeepSeek v4 Flash, a
Mixture-of-Experts model, we place them on the shared-expert MLP of every
layer. Routed experts are packed tensors that each see only a sparse
subset of tokens, whereas the shared experts run for every token, making
them the natural dense analogue of a standard MLP. Targeting all three
projections rather than the down projection alone also gives each layer
three times the adapter surface. This is a small change from prior work
with larger consequences, as it is much of what lets the Internalizer
target larger and more structurally varied backbones while unlocking
more fine-tuning capacity downstream.

In summary: the Internalizer reads a context through the frozen base model,
compresses each (\texttt{layer}, \texttt{chunk}) of activations into per-slot latents, decodes
these, together with a generated context-independent bias block, into
LoRA rows, and concatenates the blocks into one standard adapter. All of the
heavy computation lives in shared, model-agnostic weights, and only thin
entry and exit layers touch any particular base model. 

The next section describes how this system is trained end-to-end so 
that the adapters it generates actually internalize their contexts.

\section{Pre-training}
\label{sec:pretraining}

We currently train the Internalizer in a single generalised pre-training
phase, deferring task-specific fine-tuning to later work. Pre-training
teaches the hypernetwork one thing, to encode as much of a context as
possible into the adapter. Which tokens of a context matter is
task-dependent, and at pre-training time there is no task, so we treat
every token as worth internalizing. A subsequent fine-tuning phase can
then teach the system to over-weight the tokens a task cares about
(e.g.\ facts, names, figures) and to disregard the rest, starting from
an adapter that already carries them all.

Pre-training is organised as a curriculum of short training stages, each
a single epoch over a fresh slice of the corpus, with the document
length cap raised as the curriculum progresses. We first train the shared
trunk from scratch against a roster of small base models over twelve
stages, then port the warmed hypernetwork to DeepSeek v4 Flash over a
further four. The task, data recipe, and objective described below
are common to every stage. Two ideas carry this section, an error-gated
loss that focuses gradient on the tokens not yet internalized, and a
reverse distillation schedule that does the bulk of the learning against
cheap small models. The last two subsections describe the two
phases of the curriculum in turn.

\subsection{The regurgitation task}

Our pre-training task is regurgitation, in which the base model must
reproduce a document verbatim given only the adapter generated from it.
Each training sample is a (\texttt{context}, \texttt{input}, \texttt{output}) triplet in which the
context and the output are both the document, and the input is the fixed
instruction ``Reproduce the document.''. The instruction is one constant
string across the entire corpus, so it carries no information about any
particular document, and everything the model reproduces must flow
through the generated adapter. This fixed input is a pre-training
device only and is replaced by real task inputs at fine-tuning time.

\subsection{Data}

Documents come from the finepdfs corpus~\citep{kydlicek2025finepdfs}, filtered to English
documents with a language-identification confidence above 0.9. A small
fraction of documents is held out for validation, and each training
stage draws on fresh corpus shards disjoint from every earlier stage.
Each document is truncated to the stage's length cap and split into
near-equal contiguous chunks of at most 512 tokens. Every chunk is
wrapped in the base model's user-turn chat template, so the encoder
reads it on-distribution, and is prefixed with a position tag
\texttt{[doc\_id:chunk\_idx]} carrying a per-document identifier and the
chunk's index, so each per-chunk LoRA block has information of which part of which
document it encodes. We include these tags because we expect chunk
attribution, knowing which document and position a piece of
internalized content came from, to be a critical capability for
downstream fine-tuning tasks such as citation and multi-document use.
The tags are therefore not a training-only device and are kept in place
at evaluation and inference time too.

\subsection{Training loop}

Each step runs the architecture of Section~\ref{sec:arch} end to end.
The frozen base model encodes the tagged chunks, and the hypernetwork
maps the resulting per-layer activations to one rank-16 LoRA block per
chunk. These blocks are concatenated along the rank axis with the
generated bias block, and the assembled adapter is installed in the same
frozen base model. The modulated model then processes
the sample formatted as a chat, with the instruction as the user turn
and the document as the assistant turn. Supervision is teacher-forced
next-token cross-entropy computed over the assistant span only. The
instruction and chat scaffolding are masked out of the loss. The
gradient flows through the base model's frozen weights and the assembled
adapter back into
the hypernetwork, whose weights are the only ones ever updated. All stages
use a constant learning rate, between $10^{-5}$ and $4\times10^{-5}$
depending on the stage.

\subsection{Error-gated loss}

Tokens are not equally hard to internalize, as common, predictable
tokens are absorbed quickly while the hardest ones, such as names and
numbers, are often exactly those a downstream task depends on.
Because easy tokens vastly outnumber hard ones, plain cross-entropy
spends most of its gradient mass polishing tokens the adapter has
already internalized. We therefore add an error-gated term. With
$\mathcal{L}$ the mean cross-entropy over the output span and
$\mathcal{L}_{\text{err}}$ the same loss masked to tokens where the
modulated model's argmax prediction is wrong, we train on
\begin{equation}
\mathcal{L}_{\text{train}}
  = \mathcal{L} + \lambda\,\mathcal{L}_{\text{err}},
\end{equation}
so tokens not yet internalized receive $(1{+}\lambda)\times$ the
baseline gradient weight while learned tokens keep weight 1. Training on
the gated term alone proved unstable, so the plain cross-entropy is kept
as an anchor. We introduce the term partway through pre-training and use
$\lambda$ between 0.2 (early stages) and 1.5 (the final DeepSeek v4
Flash stages). Validation loss is always reported ungated, so it stays
comparable across stages.

\subsection{Warm-starting on small models}

The early gradient steps of the hypernetwork do not require an expensive
base model, so we use a strategy we call reverse distillation, named in
contrast to conventional distillation, which transfers capability from a
large model into a small one. Here the hypernetwork does the bulk of its
learning cheaply against small base models before being applied to a much
larger one. Exploiting the trunk's portability, we run the early stages of
pre-training against a roster of five small open models of 1--3B
parameters (Qwen3.5-2B, Gemma~4~E2B, MiniCPM5-1B, Nanbeige4.1-3B, and
Granite-4.1-3B) sharing the single trunk, with micro-steps cycling the
members in lockstep so that every optimizer step blends gradients from
all of them. The twelve stages of this phase ratchet the document cap from
64 tokens up to 4096, and with it the chunk count from one to eight,
alternating cheap single-model stages that train only the trunk against
frozen adapters with full-roster stages that train everything. The
stage boundaries were chosen ad hoc as training progressed, and the
schedule could likely be simplified. The warm start itself, rather than
the precise schedule, is what carries the result. A trunk pre-trained
this way converges far faster when ported to a large model than a
hypernetwork trained against that model from scratch, as shown in
Section~\ref{sec:portability}. This substantially reduces training
cost. Each step against DeepSeek v4 Flash is far slower than a step
against the roster of small models, so the large model still dominates the absolute
GPU hours of pre-training, but the warm start front-loads the coarse
learning onto cheap models and leaves only finer-grained refinement to
the expensive one, reducing the cost of training to the point where training on
state-of-the-art LLMs becomes feasible.

\subsection{Porting to DeepSeek v4 Flash}

We then port the warmed hypernetwork to DeepSeek v4 Flash 0731 (284B
parameters). The model is released as an fp8 checkpoint, which our
bf16 training stack cannot load directly, so we dequantize it to bf16
offline before training. The porting stages, on the same data
recipe, cold-start the new
model-specific entry and exit layers against the frozen trunk on
512-token single-chunk documents, continue them on 4096-token chunked
documents with the trunk still frozen, and finally train everything
unfrozen. This porting phase ran on 64 H200 GPUs (8
nodes of 8, DeepSpeed ZeRO-3) for roughly three days.

\section{Evaluation Method}
\label{sec:eval}

Pre-training aims to make the generated adapter carry the content of
its context document, so our evaluation measures how much of a
document can be recovered from the generated adapter alone.

We take three test sets of plain-text documents and truncate each
document to a range of context lengths. For every document and length,
our final trained Internalizer generates an adapter, which is
applied to DeepSeek v4 Flash 0731. We measure how accurately the
modulated model reproduces each document from the adapter alone, and
compare against the unmodified base model's plain language modelling
of the same tokens.

This evaluation quantifies the extent to which context is internalized
into adapters at each context length.

The remainder of this section describes the test datasets, the adapter
generation method, and the evaluation metrics and procedure.

\subsection{Data}
\begin{table*}[t]
\centering
\small
\begin{tabular}{rrrcccccc}
\toprule
& & & \multicolumn{2}{c}{Perplexity} & \multicolumn{2}{c}{Top-1 (\%)} & \multicolumn{2}{c}{Top-5 (\%)} \\
\cmidrule(lr){4-5} \cmidrule(lr){6-7} \cmidrule(lr){8-9}
Length & Docs & Rank & Adapter & Base & Adapter & Base & Adapter & Base \\
\midrule
\multicolumn{9}{l}{\emph{FinePDFs (train distribution)}} \\
256 & 100 & 32 & \textbf{1.27} & 7.85 & \textbf{92.6} & 57.5 & \textbf{99.5} & 79.3 \\
512 & 87 & 32 & \textbf{1.55} & 6.89 & \textbf{86.2} & 59.1 & \textbf{98.7} & 80.6 \\
1024 & 67 & 48 & \textbf{1.55} & 5.99 & \textbf{86.4} & 61.3 & \textbf{98.7} & 82.1 \\
2048 & 49 & 80 & \textbf{1.58} & 5.58 & \textbf{85.8} & 61.8 & \textbf{98.6} & 83.0 \\
4096 & 41 & 144 & \textbf{1.97} & 5.58 & \textbf{80.2} & 61.7 & \textbf{96.6} & 82.8 \\
8192$^{\dagger}$ & 24 & 272 & \textbf{3.82} & 4.81 & \textbf{67.6} & 64.5 & \textbf{87.7} & 84.7 \\
16384$^{\dagger}$ & 17 & 528 & 2724 & \textbf{4.62} & 24.6 & \textbf{65.4} & 38.2 & \textbf{85.0} \\
\midrule
\multicolumn{9}{l}{\emph{FinePDFs (long)}} \\
256 & 32 & 32 & \textbf{1.25} & 4.85 & \textbf{93.1} & 67.8 & \textbf{99.4} & 84.9 \\
512 & 32 & 32 & \textbf{1.45} & 4.66 & \textbf{88.3} & 67.7 & \textbf{98.8} & 85.7 \\
1024 & 32 & 48 & \textbf{1.49} & 4.76 & \textbf{87.3} & 65.7 & \textbf{98.8} & 85.1 \\
2048 & 32 & 80 & \textbf{1.52} & 4.85 & \textbf{86.9} & 64.6 & \textbf{98.7} & 84.8 \\
4096 & 32 & 144 & \textbf{1.82} & 4.81 & \textbf{82.0} & 64.6 & \textbf{97.2} & 84.9 \\
8192$^{\dagger}$ & 32 & 272 & \textbf{3.56} & 4.44 & \textbf{69.0} & 66.2 & \textbf{88.5} & 85.7 \\
16384$^{\dagger}$ & 32 & 528 & 2276 & \textbf{4.10} & 25.4 & \textbf{67.6} & 39.3 & \textbf{86.4} \\
\midrule
\multicolumn{9}{l}{\emph{PG-19}} \\
256 & 100 & 32 & \textbf{1.58} & 5.05 & \textbf{88.0} & 68.8 & \textbf{97.8} & 86.0 \\
512 & 100 & 32 & \textbf{1.93} & 5.58 & \textbf{81.8} & 65.6 & \textbf{96.5} & 84.5 \\
1024 & 100 & 48 & \textbf{1.93} & 5.93 & \textbf{80.9} & 62.9 & \textbf{96.7} & 83.5 \\
2048 & 100 & 80 & \textbf{1.97} & 5.99 & \textbf{80.0} & 61.2 & \textbf{96.8} & 83.1 \\
4096 & 99 & 144 & \textbf{2.48} & 5.99 & \textbf{74.1} & 60.0 & \textbf{94.4} & 82.9 \\
8192$^{\dagger}$ & 95 & 272 & \textbf{5.53} & 6.05 & \textbf{59.3} & 59.0 & \textbf{83.6} & 82.6 \\
16384$^{\dagger}$ & 91 & 528 & 1978 & \textbf{5.81} & 22.0 & \textbf{59.2} & 37.4 & \textbf{83.0} \\
\bottomrule
\end{tabular}
\caption{Document reconstruction by document set and context length, comparing
Adapter-only regurgitation against the unmodified base model. 
\emph{Rank} is the adapter's combined
LoRA rank, $16(k{+}1)$ for $k$ chunks. Accuracies are means over the
row's documents, perplexity is the exponential of the mean per-document
cross-entropy, and the better value is in bold. $^{\dagger}$Beyond the 4096-token,
eight-chunk training cap, so the hypernetwork generates adapters at
chunk counts it never saw in training.}
\label{tab:reconstruction}
\end{table*}

We evaluate on three document sets.

\textbf{FinePDFs (train distribution)} is 100 documents drawn from an
\texttt{eng\_Latn} test-split shard of finepdfs~\citep{kydlicek2025finepdfs} that no training
stage touched, under the same language filter as pre-training.

\textbf{FinePDFs (long)} draws from the same shard but keeps only
documents that reach 16384 tokens, of which the shard yielded 32.

\textbf{PG-19} is the 100 PG-19 test books~\citep{rae2020compressive}, out-of-domain
public-domain narrative prose rather than text extracted from PDFs,
unseen by every training stage.

Each document is evaluated at seven length buckets of 256, 512, 1024,
2048, 4096, 8192 and 16384 tokens. A document is truncated to the
largest bucket it reaches and evaluated at every bucket at or below it,
so the same document contributes a point at each length. The long
buckets of the FinePDFs train distribution therefore thin out as the
training length distribution does, and the document counts in
Table~\ref{tab:reconstruction} reflect that. 

In total this is 232 documents and 1294 (\texttt{document}, \texttt{length}) evaluations.

\subsection{Adapter generation}

Adapters are generated exactly as in training. A document is truncated
to the target length, then chunked, tagged, and encoded as in
Section~\ref{sec:pretraining}, and our final trained hypernetwork maps
this to one rank-16 block per chunk, giving a standard LoRA adapter of
rank $16(k{+}1)$ once the bias block is appended.

\subsection{Metrics and control}

We score each document in the regurgitation format of
Section~\ref{sec:pretraining}, teacher-forced over the assistant span.
We evaluate these adapters using the following metrics.

\begin{itemize}
  \item \textbf{Perplexity (PPL).} The exponential of the mean
    teacher-forced next-token cross-entropy over the document span.
    Cross-entropy is the pre-training objective itself, and its
    exponential reads as the effective number of tokens the model is
    choosing between at each position, so this measures internalization
    on the scale the hypernetwork was optimised for while staying
    interpretable.
  \item \textbf{Top-1 accuracy.} The fraction of tokens whose argmax
    prediction matches the true token. It is the closest teacher-forced
    proxy for how much of the document a greedy decode would reproduce
    verbatim.
  \item \textbf{Top-5 accuracy.} The fraction of tokens whose true token
    is among the five highest-probability predictions. Crediting
    near-misses distinguishes tokens the adapter carries but narrowly
    misranks from tokens it failed to internalize at all.
\end{itemize}

The control is the same document scored by the unmodified base model as
plain language modelling, with no adapter and no chat template, just the
raw document tokens.\footnote{The two supervised spans are not quite
identical. The adapter setting also supervises the chunk tags and the
turn close, some 1.5\% more tokens.} Both settings are teacher-forced, so the only difference between
them is whether the document was internalized. Base-model accuracy is
how much of a document a 284B-parameter model can predict having never
seen it, and any margin above that line is content the adapter carried
into the weights rather than content the model could have guessed.

\section{Results}
\label{sec:results}

\begin{figure*}[t]
    \centering
    \includegraphics[width=\textwidth]{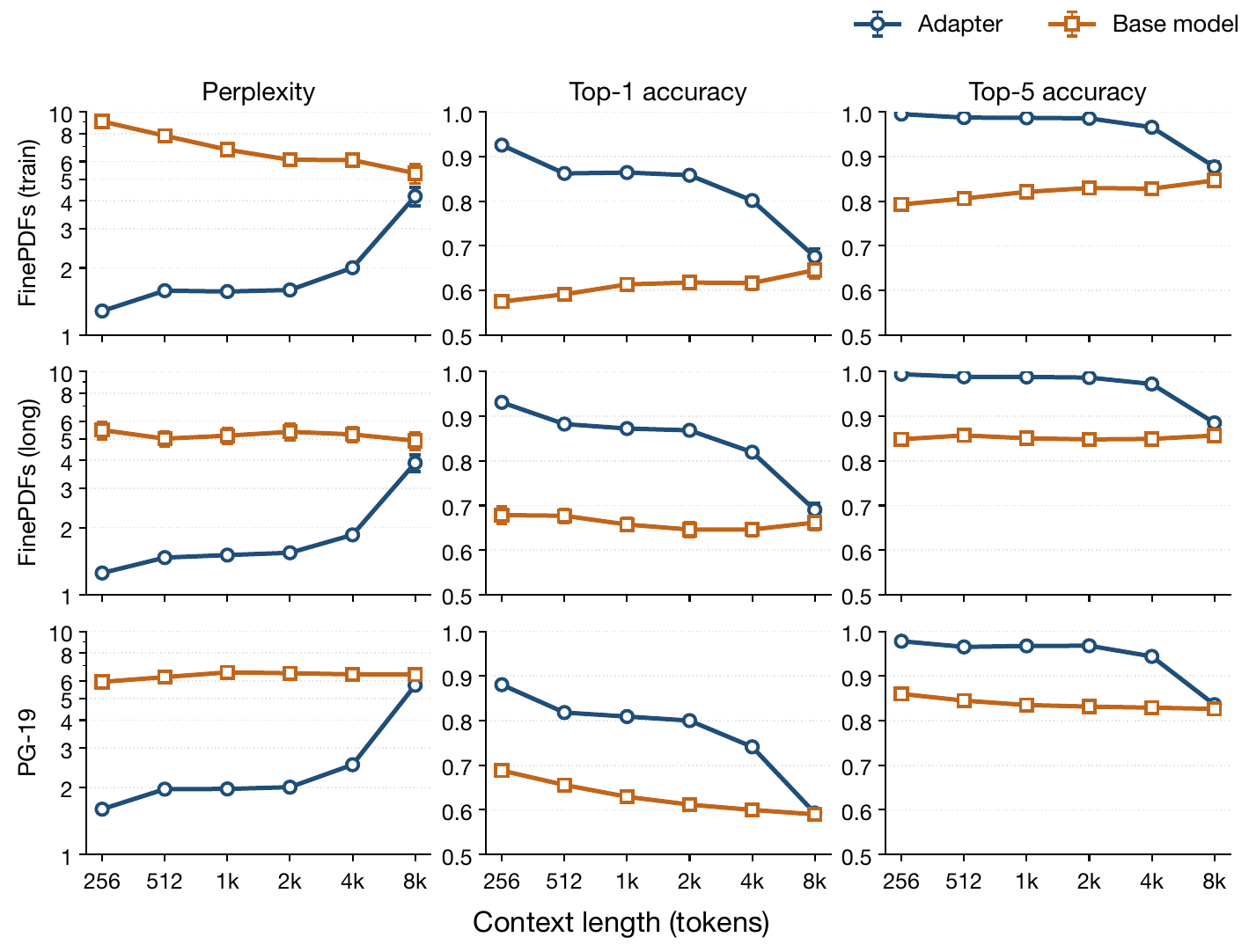}
    \caption{Adapter versus base model across context lengths, one
    document set per row. FinePDFs (train distribution) covers 100
    unique documents (24--100 per context length, 368 evaluations),
    FinePDFs (long) holds the same 32 documents at every context length
    (192 evaluations), and PG-19, an out-of-domain set, covers 100
    unique documents (95--100 per context length, 594 evaluations).
    Error bars show the standard error of the mean. The 16384-token
    bucket is omitted, as its adapter perplexities would dominate the
    axis. It is discussed in Section~\ref{sec:eval-extrapolation}.}
    \label{fig:reconstruction_curves}
\end{figure*}

Our results show that the hypernetwork internalizes documents
accurately at the lengths it was trained on, but this does not
generalize to longer ones.
Table~\ref{tab:reconstruction} gives the full numbers and
Figure~\ref{fig:reconstruction_curves} the trends.

\subsection{Within the trained length}
\label{sec:eval-trained}

Within the trained regime of documents up to 4096 tokens and eight
chunks, the hypernetwork internalizes nearly all of each document's content
into the adapter. Macro-averaged over the five buckets
up to that length and the three document sets, the adapter reaches
84.9\% top-1 and 97.8\% top-5 accuracy against 63.4\% and 83.5\% for the
base model, at a perplexity of 1.7 against 5.6. Under teacher forcing,
the vast majority of tokens in a document the model has never seen are
predicted correctly from the adapter weights alone, with nothing in the
context window but a three-word instruction.

Base-model accuracy is largely flat in length on all three sets,
while adapter accuracy varies with it. The margin between the two
therefore reflects how much of the document reached the adapter rather
than how predictable the text is.

Internalization accuracy is also only weakly affected by document
length within this regime. From 256 to 2048 tokens, an eightfold
increase in content, top-1 accuracy falls by only six points on
FinePDFs (long). The adapter's rank grows with the chunk count, so its
capacity scales with the amount of content it must store. The first
clear drop in accuracy appears at 4096 tokens, where the chunk count
reaches the training cap.

PG-19 shows the lowest internalization accuracy throughout, at 81.0\%
macro top-1 up to 4096 tokens against 86.2\% and 87.5\% for the two
FinePDFs sets. Its text is not intrinsically less predictable, as the
base model actually scores slightly higher on PG-19, so the gap is in
internalization rather than in the text. Narrative prose has none of
the layout regularity of extracted PDF text and is out of domain for
every training stage. The adapter's margin over the base model narrows
accordingly, from 25.9 points on the FinePDFs train distribution to
17.3 on PG-19.

\subsection{Beyond the trained length}
\label{sec:eval-extrapolation}

The last two rows of each block in Table~\ref{tab:reconstruction} are
extrapolation. At 8192 and 16384 tokens the hypernetwork must produce 16
and 32 chunks against a maximum of 8 in training, with combined ranks
of 272 and 528 rather than 144.

At 8192 tokens adapter accuracy has fallen back to roughly the base
model's level, with top-1 margins of 0.3 to 3.1 points across the three
sets, so little of the document is reaching the adapter. At 16384
tokens adapter accuracy drops well below the base model's, to 22--25\%
top-1, with perplexities in the thousands, indicating significant
divergence, which is why we omit the bucket from the figures.

Internalization accuracy outside the trained length range is therefore
low. Within that range of up to about 4096 tokens per adapter, a single
hypernetwork forward pass yields roughly 22 points of top-1 accuracy
over the 284B-parameter base model on documents it has never seen,
with the adapter as the model's only source of the document. That
knowledge is delivered as weights, generated in one pass and reusable
across every query, to a model of a class whose accuracy is already
high enough for real deployment. Our aim is an accuracy mode that adds
that knowledge on top of the context held in the model's context
window, rather than only a speed mode that replaces it, so that the
adapter raises accuracy above what in-context learning or standard
LoRA training alone achieves.

\subsection{Portability}
\label{sec:portability}

\begin{figure*}[t]
    \centering
    \includegraphics[width=0.67\textwidth]{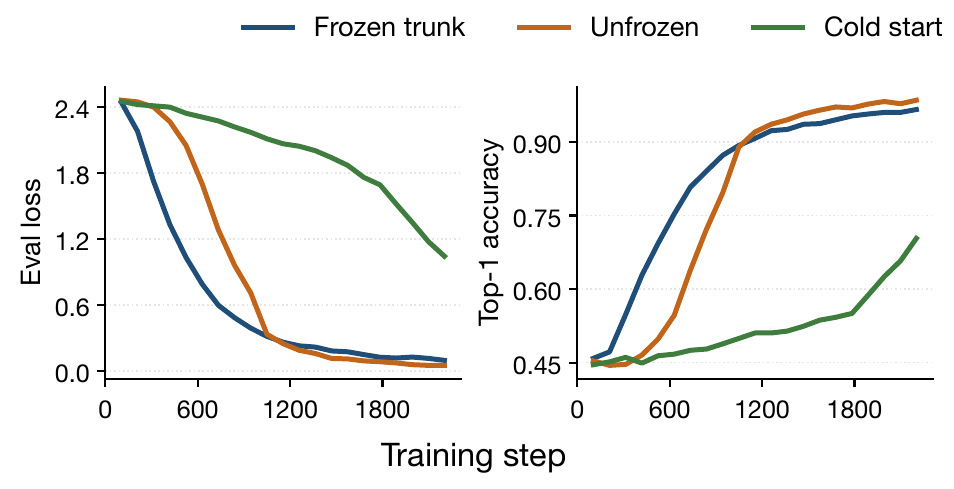}
    \caption{Evaluation loss and top-1 accuracy while porting to
    DeepSeek v4 Flash. The trunk-frozen and fully unfrozen runs start
    from the warmed shared trunk, while the cold-start run trains its
    hypernetwork from scratch.}
    \label{fig:portability}
\end{figure*}

Finally, we run a small experiment to isolate how much of the ported
hypernetwork's performance comes from the shared trunk rather than from
the thin model-specific entry and exit layers, and with it to measure
the value of reverse distillation. The experiment uses 64-token
single-chunk documents, where training is cheap enough to repeat. We
first pre-train a hypernetwork from a cold start against the five-model
small roster of Section~\ref{sec:pretraining}, which is fast, and keep
its shared trunk. We then train three hypernetworks against DeepSeek v4
Flash on the same data.

\begin{itemize}
  \item \textbf{Frozen trunk.} The warmed trunk is kept frozen and only
    cold-started entry and exit layers for DeepSeek v4 Flash are
    trained.
  \item \textbf{Unfrozen.} The warmed trunk and cold-started entry and
    exit layers are all trained.
  \item \textbf{Cold start.} The entire hypernetwork is trained from
    scratch against DeepSeek v4 Flash.
\end{itemize}

Figure~\ref{fig:portability} shows evaluation loss and top-1 accuracy
over the course of training. Both warm-started runs pass 90\% top-1
accuracy within the training budget, while the cold start does not reach
above 75\% at the same step count and is still improving slowly. The
frozen-trunk run gains accuracy in the fewest steps, since its only
trainable weights are the thin per-model maps, and the fully unfrozen
run finishes marginally ahead of it.

The two warm-started curves cross at roughly step 1000. Our reading is
that freezing the trunk helps early because it forces all gradient into
the randomly initialised per-model maps, which must first reach a
plausibly trained state, and then becomes a brake because those thin
maps have far less capacity than the trunk. The effect may also be
specific to this cheap 64-token setting. At the 4096-token scale of the
main porting run, a short frozen phase made little difference to loss
and we moved to unfrozen training quickly, though that may simply
reflect the scale of that run.

That a frozen trunk trained only against 1--3B models can be fitted to
a 284B-parameter model through thin entry and exit layers alone shows
that most of the ported performance comes from the shared trunk. This
is the evidence behind our reverse distillation strategy, in which the
bulk of the learning happens quickly and cheaply against small models
and the expensive training against the large model only has to fit the
thin model-specific layers.

\section{Future work}
\label{sec:futurework}

This work pre-trains the Internalizer and evaluates it on
reconstruction. We plan to build on it in the following directions.

\begin{itemize}

  \item \textbf{Task finetuning.} Reconstruction measures how much of a
        document reaches the adapter, but it is not itself an
        application. The next step is to keep the same training loop
        and swap the reconstruction prompt for real questions,
        instructions, and answers from tasks such as question answering
        over documents and tables, code generation against an
        internalized repository, and instruction following grounded in
        internalized manuals and policies. Every such finetune branches cheaply from
        the same pre-trained hypernetwork.

  \item \textbf{Full-context finetuning.} Internalization pre-training
        removes the context from the window, but finetuning need not.
        We plan to finetune with the same context both fed to the
        hypernetwork and kept in the window.
        The adapter then no longer stands in for the context. It becomes
        a context-conditioned specialization of the task, learning the
        task itself while adjusting to the sub-type of task the context
        implies. This is a superset of standard LoRA training, where the
        adapter learns the task from a large corpus, as a LoRA would,
        and additionally fits itself to each specific context, combining
        generalization with specialization.

  \item \textbf{Longer context lengths.} Our results show that
        internalization accuracy does not extend beyond the trained
        length range, so a longer useful range has to be trained for
        rather than expected from extrapolation. We plan to raise the
        training cap well beyond eight chunks and measure how accuracy
        and training cost scale with it.

  \item \textbf{Transfer to other and larger models.} Prior work
        suggests that model representations become more aligned as
        models grow~\citep{huh2024platonic}, at least in their local
        neighborhood structure~\citep{groger2026aristotelian}. We
        therefore expect the trunk trained against DeepSeek v4 Flash to
        port to other large models at a much lower cost compared to
        training them from scratch, just as the small-roster trunk
        ported to DeepSeek v4 Flash. The natural targets are
        trillion-parameter models such as Qwen3.8-2.4T-A95B and Kimi
        K3, where the savings from reverse distillation are largest.

  \item \textbf{Harness-specific training.} Deployed models run inside
        a harness of tools, prompting conventions, and output formats.
        Finetuning the Internalizer against a specific harness would
        let its adapters absorb those conventions, and this is the
        route we expect commercial applications to take.

  \item \textbf{Self-improvement of adapters through reflection.} Because
        an adapter is a cheap function of its context, a deployed adapter
        can be improved without any weight training. During down-time, a
        strong but slow model could review the adapter's mistakes or mine
        real user interactions, append what it learns to the context, and
        regenerate the adapter, yielding a smarter adapter without the
        cost or instability of actual training. Deployment then becomes
        a hybrid mode, serving with the current adapter while reflective
        refinement improves the next one offline.

  \item \textbf{N-gram embedding targets.} The Qwen 3.8 Next
        architecture~\citep{qwen2026next} adds n-gram embedding tables
        that can be offloaded out of GPU memory. Generating adapter components that target
        this embedding space, or a task-relevant subset of it, could
        pair internalization with that offloading for further inference
        savings.
\end{itemize}

\section{Conclusion}
\label{sec:summary}

We presented the Internalizer, a state-of-the-art, portable
Context-to-Parameter Mapping hypernetwork that converts a document
into a LoRA adapter for a frozen base model. Its trunk is shared across base models, with only thin
entry and exit layers specific to each, and its target is DeepSeek v4
Flash at 284B parameters, twenty times larger than any prior
target in this area.

Pre-trained on a regurgitation task, the hypernetwork internalizes
unseen documents of up to 4096 tokens at 84.9\% top-1 and 97.8\% top-5
teacher-forced accuracy, roughly 22 points of top-1 over the base
model, with nothing in the context window.

These results point to task-specific context becoming something a
state-of-the-art model computes once as a LoRA and keeps. Such adapters
were previously available only to small models, whose base accuracy
left little to build on. At this scale the base accuracy is high enough
for real deployment, and the adapter can lift it further, alone in a
speed mode or alongside the context in the window in an accuracy mode.
In either mode the adapter is conditioned on the context itself and 
carries both information of the task and the context, which
a standard LoRA adapter cannot.

Documents, schemas, or codebases can be internalized ahead
of time into compact adapters, swapped in per request, and reused
across queries, and the portable trunk lets this machinery follow
base-model upgrades at the cost of refitting the thin entry and exit
layers.

We do not present these findings as a finished model, but as a
pre-trained basis on which to build. The immediate next step is to finetune
it against real tasks and measure the accuracy its adapters add in
deployment.

\section*{Contributions}
Peter Devine was the lead researcher on this project, having co-created the concept, designed and run the experiments and evaluations, and written this paper.
Nick Ryan and Ben Sirb both consulted on this work, gave input to research direction, and reviewed various drafts of this paper before release.
Alex Chiocchi supervised this work, co-created the concept, consulted on its direction, and also reviewed this paper before release.

\bibliography{custom}

\end{document}